\pdfoutput=1

\documentclass[11pt]{article}

\usepackage[]{acl}

\usepackage{times}
\usepackage{latexsym}

\usepackage[T1]{fontenc}

\usepackage[utf8]{inputenc}

\usepackage{microtype}

\usepackage{inconsolata}

\usepackage{listings}
\usepackage{pgfplots}
\pgfplotsset{compat=1.18}
\usepackage{cleveref}
\usepackage{graphicx}
\usepackage{soul}
\usepackage{booktabs} 
\usepackage{multirow}
\usepackage{fontawesome5}
\newcommand{\code}[1]{\texttt{#1}}
\newcommand{\datasetname}{\textsc{PuzzleVQA}}

\title{\datasetname: Diagnosing Multimodal Reasoning \\Challenges Through Abstract Patterns}

\title{
\textsc{Are Language Models Puzzle Prodigies?}
\\ Diagnosing Multimodal Reasoning Challenges Through Abstract Patterns}
\title{
\textsc{Are Language Models Puzzle Whiz?}
\\ Diagnosing Multimodal Reasoning Challenges Through Abstract Patterns}

\title{
\textsc{Are Language Models Puzzle Amateurs?}
\\ Diagnosing Multimodal Reasoning Challenges With Abstract Patterns}

\title{\datasetname{}:  Diagnosing Multimodal Reasoning Challenges of
Language Models with Abstract Visual Patterns}

\author{Yew Ken Chia$^{1,2}$\thanks{\; Yew Ken Chia is under the Joint Ph.D. Program between DAMO Academy and the Singapore University of Technology and Design.}, Vernon Toh Yan Han$^1$, Deepanway Ghosal$^1$,\\ \textbf{Lidong Bing$^2$, Soujanya Poria$^1$} \\\\
$^1$ Singapore University of Technology and Design, 
$^2$ DAMO Academy, Alibaba Group, Singapore\\
}
\begin{document}
\maketitle

\begin{minipage}[t]{2\linewidth}
\vspace{-1.5cm}
  \centering
  \faGithub: \url{https://github.com/declare-lab/LLM-PuzzleTest} \\
  \faGlobe : \url{https://puzzlevqa.github.io/}
\vspace{0.5cm}
\end{minipage}

\begin{abstract}
Large multimodal models extend the impressive capabilities of large language models by integrating multimodal understanding abilities.
However, it is not clear how they can emulate the general intelligence and reasoning ability of humans.
As recognizing patterns and abstracting concepts are key to general intelligence, we introduce \datasetname{}, a collection of 2000 puzzle instances based on abstract patterns.
With this dataset, we evaluate large multimodal models with abstract patterns based on fundamental concepts, including colors, numbers, sizes, and shapes.
Through our experiments on state-of-the-art large multimodal models, we find that they are not able to generalize well to simple abstract patterns.
Notably, GPT-4V achieves a score of 46.4\% on single-concept puzzles, which shows that state-of-the-art models struggle on our dataset.
To diagnose the reasoning challenges in large multimodal models, we progressively guide the models with our ground truth reasoning explanations for visual perception, inductive reasoning, and deductive reasoning.
Our systematic analysis finds that the main bottlenecks of GPT-4V are weaker visual perception and inductive reasoning abilities. 
Through this work, we hope to shed light on the limitations of large multimodal models and how they can better emulate human cognitive processes in the future\footnote{Our data and code are released at \href{https://github.com/declare-lab/LLM-PuzzleTest}{https://github.com/declare-lab/LLM-PuzzleTest}.}.
\end{abstract}

\section{Introduction}


\begin{figure}[!t]
\centering
\resizebox{0.9\linewidth}{!}{
\includegraphics{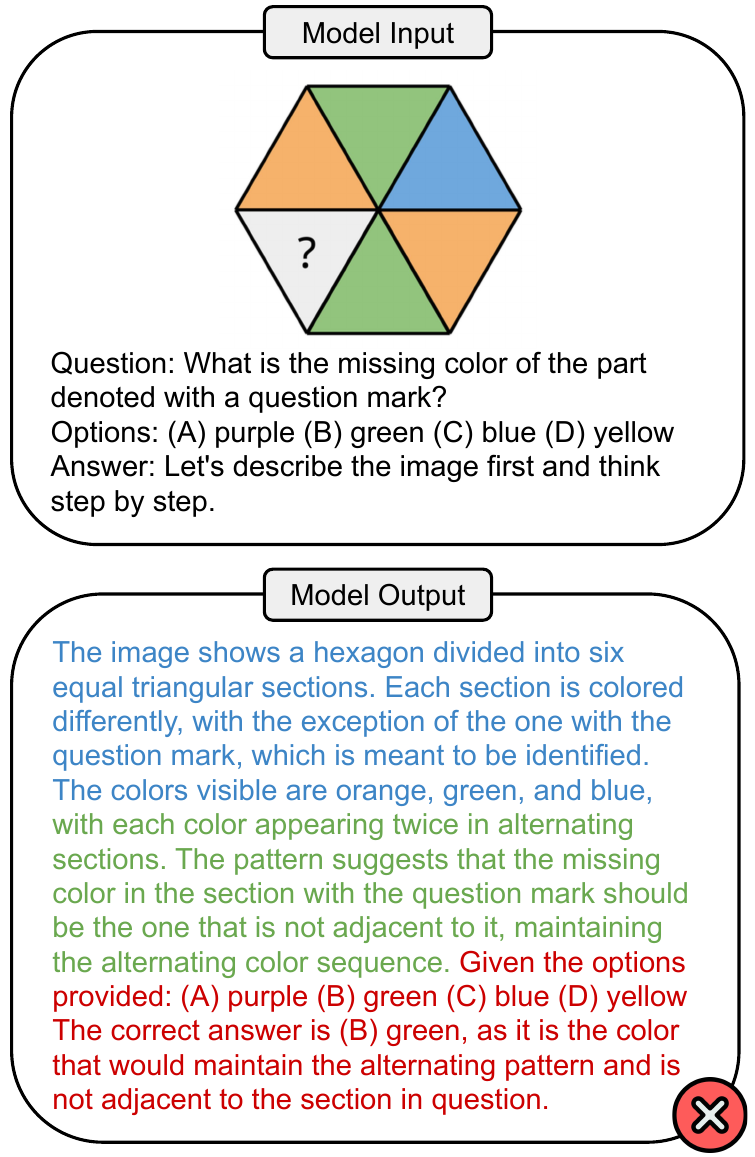}}
\caption{An example question which involves the color concept in \datasetname{}, and an incorrect answer from GPT-4V. There are generally three stages that can be observed in the solving process: visual perception (blue), inductive reasoning (green), and deductive reasoning (red).
Here, the visual perception was incomplete, causing a mistake during deductive reasoning.
}
\label{fig:intro}
\end{figure}
Rapid advances in large language models have demonstrated remarkable capabilities across diverse language tasks and applications \cite{sparks, gpt3, llama2}.
To enable more general capabilities, large multimodal models were introduced by integrating large language models with multimodal understanding \cite{yue2023mmmu, dawn, gpt4vision}.
However, it is not clear how large multimodal models can emulate the general intelligence and reasoning ability of humans \cite{puzzling}.
Specifically, we aim to explore how large multimodal models can emulate cognitive processes to perceive and interpret information, extrapolate from observations to broader generalizations, and apply general principles to solve specific problems \cite{piaget1976piaget}.
Furthermore, we are interested in understanding how well the models can reason about fundamental concepts such as numbers, colors, shapes, and size \cite{eyes, visioncheck}.


As pattern recognition and and abstracting concepts are at the heart of general intelligence \cite{machinesthink, originconcepts, fluidconcepts}, we believe that abstract patterns are a suitable testbed for evaluating reasoning ability in large multimodal models.
As shown in Figure \ref{fig:intro}, abstract patterns enable us to focus on one or more abstract concepts, and decompose the multimodal reasoning process into several stages that mimic human cognitive processes \cite{piaget1976piaget}.
Firstly, the model requires visual perception to understand and interpret the abstract image in the input.
Secondly, the model requires inductive reasoning to relate the observations shown and form a hypothesis for the underlying pattern.
Thirdly, the model requires deductive reasoning to apply the general principle of the pattern to solve the specific problem at hand.
While the abstract patterns may seem simple, we surprisingly find that even advanced large multimodal models such as Gemini Pro \cite{geminipro} and GPT-4V \cite{gpt4vision} struggle to understand them.


Puzzles are problems that require ingenuity and creativity to solve, and they can serve as valuable tools for cognitive development and assessment \cite{zhang2019raven}.
Hence, we propose the \datasetname{} dataset to systematically evaluate and diagnose the reasoning challenges in large multimodal models.
Our dataset consists of diverse multimodal puzzles that focus on abstract patterns with fundamental concepts including numbers, colors, shapes, and size.
We design and automatically construct the dataset through multimodal templates, enabling us to generate large numbers of puzzles without costly human annotation \cite{dataannotator}.
To support interpretability and systematic investigation of reasoning challenges in multimodal models, we also construct the ground truth reasoning explanations for each puzzle.
Compared to existing datasets for visual question answering, \datasetname{} focuses specifically on how large multimodal models can mimic general cognitive processes such as inductive and deductive reasoning. 
As we focus on how models can generalize to novel problems, similar to fluid intelligence in humans \cite{fluid}, our dataset in the abstract domain poses challenges for existing models without requiring extensive world knowledge. 


Through our investigation of leading large multimodal models, we find that existing models are not able to generalize well to simple abstract patterns.
Notably, GPT-4V achieves a score of 46.4\% on single-concept puzzles, which shows that state-of-the-art models struggle on our dataset.
Our analysis reveals that its main bottlenecks are weaker visual perception and inductive reasoning abilities. 
Hence, our main contributions include:

\begin{enumerate}
    \item To investigate the cognitive and reasoning abilities of large multimodal models, we propose to leverage abstract patterns.
    \item We introduce \datasetname{}, an automatically generated and diverse dataset of 2000 multimodal samples with reasoning explanations.
    \item Our experiments show that even advanced large multimodal models do not generalize well to abstract patterns, and we show how to identify their reasoning bottlenecks.
\end{enumerate}

\begin{figure*}[t]
    \centering
    \includegraphics[width=0.8\textwidth]{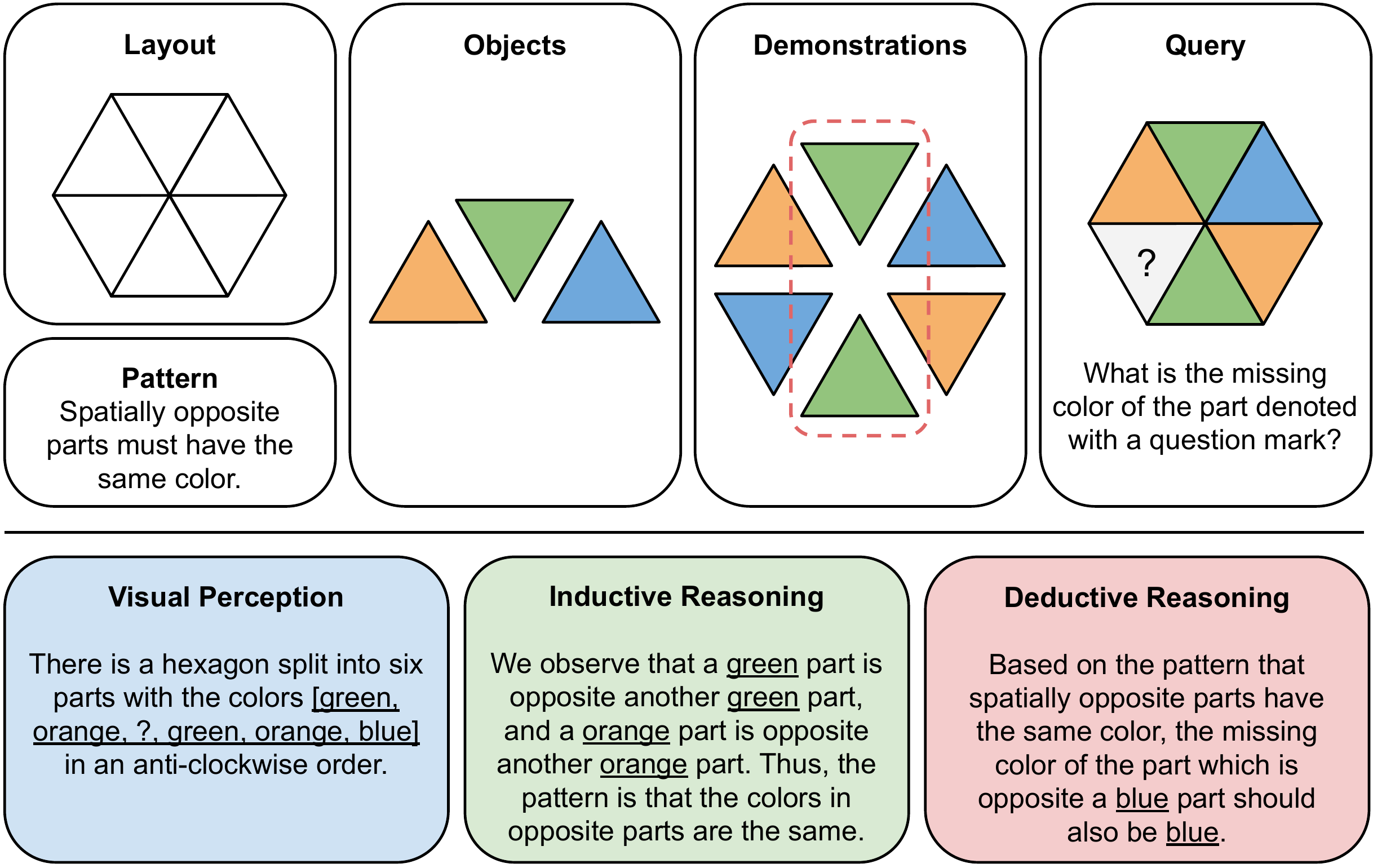}
    \caption{Illustration example of components (top) and reasoning explanations (bottom) for abstract puzzles in \datasetname{}. To construct each puzzle instance, we first define the layout and pattern of a multimodal template, and populate the template with suitable objects that demonstrate the underlying pattern. For interpretability, we also construct ground truth reasoning explanations to interpret the puzzle and explain the general solution stages.}
    \label{fig:components}
\end{figure*}

\section{Background: Cognitive Theories}
\label{sec:cognitive}
To understand how large multimodal models can better mimic human thought processes and general intelligence, we first ground our study with relevant cognitive theories.

\subsection{Fluid and Crystallized Intelligence}
The Cattell-Horn theory \cite{fluid} of cognitive abilities distinguishes between two types of intelligence: fluid intelligence, which involves the ability to solve novel problems without relying on previously acquired knowledge, and crystallized intelligence, which involves the use of knowledge, skills, and experience. 
Fluid intelligence in humans could parallel large multimodal models' ability to solve new, unseen problems through pattern recognition and problem-solving strategies. 
On the other hand, crystallized intelligence could be akin to how the models leverage accumulated world knowledge from training data to understand and interact with the world \cite{cognitivearch}.
As many works have focused on how models can leverage specialized knowledge \cite{yue2023mmmu}, we instead focus on how they may emulate fluid intelligence to solve novel problems through abstract patterns.

\subsection{Cognitive Development}
Piaget's Stages of Cognitive Development \cite{piaget1976piaget} can provide a framework for progressing from basic sensory experiences to complex abstract reasoning and problem-solving. 
While we note that the large multimodal models do not develop in the same organic and experiential manner as humans, we are guided to explore how the models can emulate different stages of cognitive abilities. 
Concretely, through abstract patterns, we can evaluate how the models perceive multimodal information, reason inductively to extrapolate from observations to broader generalizations, and apply general principles to deduce the solution for specific problems.

\paragraph{Sensorimotor.} This stage underpins visual perception, where individuals learn to coordinate sensory experiences through interactions with the environment.
To emulate this cognitive stage, we would expect models to identify simple shapes or colors but lack higher-level reasoning.
Hence, we set the foundation for later stages by exploring abstract patterns based on fundamental concepts including colors, numbers, shapes, and size.

\paragraph{Preoperational.} At this stage, individuals develop symbolic thinking, which is crucial for understanding representations in visual contexts and beginning to engage in simple, inductive reasoning processes.
Models that mimic this stage should be able to perform basic reasoning about objects or concepts, but with limited understanding of abstract relationships or performing logical operations.

\paragraph{Concrete Operational.} This stage is closely related to inductive reasoning, as individuals learn to think logically about concrete events and solve problems based on visible patterns and relationships.
We would expect models that are analogous to this stage to have the ability to draw logical conclusions from specific instances and start to apply these conclusions to solve problems.
Hence, we consider inductive reasoning as an integral part of understanding abstract patterns.

\paragraph{Formal Operational.} This stage is essential for deductive reasoning and abstract thinking, allowing individuals to hypothesize and think about theoretical scenarios, which are skills necessary for solving complex problems. 
At this stage, we would expect comparable models to effectively induce general principles or hypotheses from observations and logically deduce specific outcomes, even in abstract or novel contexts.
Thus, we consider deductive reasoning as critical to solving abstract problems.

\section{PuzzleVQA Dataset}

Despite the impressive capabilities of large multimodal models, we do not fully understand how they solve multimodal problems through reasoning.
Specifically, we focus on how well they can interpret multimodal inputs, form generalizations from observations, and apply the general principles to solve specific cases.
Furthermore, they may reason differently about fundamental concepts such as numbers, colors, shapes, and size.
Hence, we propose \datasetname{}, a diverse collection of abstract pattern puzzles to diagnose reasoning challenges in multimodal models.
The dataset is automatically constructed through multimodal templates, and includes reasoning explanations for interpretability.

\subsection{Puzzle Components}
\label{sec:components}
As shown in Figure \ref{fig:components}, each puzzle in our dataset is formulated with the following main components:  

\begin{enumerate}
    \item \textbf{Objects:} The conceptual elements that interact within the puzzle, such as numbers, colors, shapes, and size.
    \item \textbf{Layout:} The spatial arrangement of objects that provides visual context.
    \item \textbf{Pattern:} The relationship that governs the interaction amongst objects. For example, a pattern may be that spatially opposite parts must have the same color.
    \item \textbf{Demonstrations:} Multiple instances of interacting objects that collectively represent the underlying pattern. Without demonstrations, the pattern would become ambiguous.
    \item \textbf{Query:} The natural language question that directs the multimodal model how to solve the puzzle by determining the missing object.
\end{enumerate}

\subsection{Design Considerations}
We have three main design considerations for each puzzle in our dataset:

\paragraph{Simplicity.} As the focus is on evaluating how large multimodal models reason about fundamental abstract concepts, we do not deliberately make the puzzles more complex than necessary. We also aim to make the underlying patterns straightforward, without requiring extensive world knowledge or advanced theories.

\paragraph{Correctness.} To avoid potentially noisy annotations, we use an automatic approach with multimodal templates to generate each puzzle. For instance, given a visual layout and pattern of a template, we can automatically populate the template with the appropriate objects that demonstrate the pattern.
As each puzzle instance is created based on the specific rules in the template, we can ensure that they do not contain annotation mistakes. 

\paragraph{Diversity.} To investigate the multimodal reasoning capabilities across diverse abstract concepts, we construct puzzles based on four main concepts: numbers, colors, shapes, and size. Furthermore, to evaluate how well the models can relate to multiple concepts, we design both single-concept and dual-concept puzzles. 


\begin{figure*}[t]
    \centering
    \includegraphics[width=0.9\textwidth]{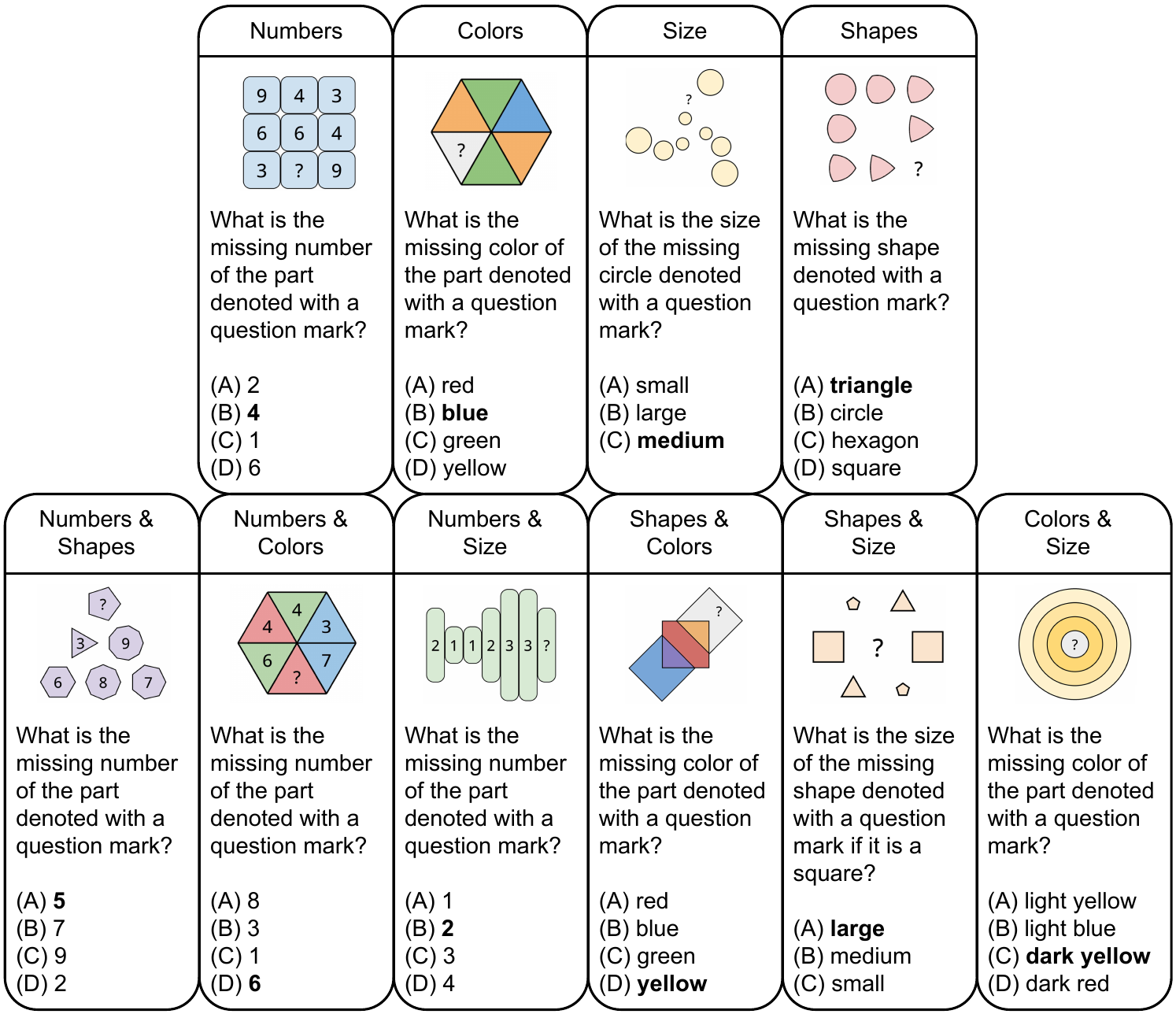}
    \caption{Taxonomy of abstract puzzles in \datasetname{} with sample questions, based on fundamental concepts such as colors and size. To enhance diversity, we introduce both single-concept and dual-concept puzzles.}
    \label{fig:ontology}
\end{figure*}

\subsection{Puzzle Construction}
\paragraph{Multimodal Templates.}
To construct each abstract puzzle, we leverage multimodal templates based on fundamental concepts including numbers, colors, shapes, and size.
Following the formulation and design considerations previously discussed, we first define the layout and abstract pattern for the puzzle.
Each template can be randomly populated with the specific objects to represent the underlying pattern through demonstrations, forming a specific puzzle instance.
For example, to construct a color-based puzzle instance shown in Figure \ref{fig:components}, we focus on the concept of colors and define the layout as a hexagonal arrangement of six parts, with the abstract pattern that spatially opposite parts must have the same color.
Thereafter, the template can be randomly populated to satisfy the pattern with colors from a predefined list of possible colors.
Lastly, the query is constructed based on the fundamental concepts in the abstract pattern.
To demonstrate our puzzle generation pipeline, we include a detailed implementation in Appendix \ref{sec:code_snippet}, based on the puzzle in Figure \ref{fig:intro}.

\paragraph{Reasoning Explanations.}
To ensure that each abstract puzzle can be easily understood, we also construct reasoning explanations based on the three problem solving stages: image descriptions for visual perception, pattern explanations for inductive reasoning, and deductive reasoning steps.
Specifically, we leverage textual templates that can be populated with details from the specific puzzle instance, as shown in Figure \ref{fig:components}.
In our experiments in Section \ref{sec:results}, this enables us to identify reasoning bottlenecks by progressively providing the explanations of each stage to the model.

\subsection{Multiple-Choice Format}
While we use straightforward objects in our puzzles, there may be a degree of ambiguity in the answer regarding specific colors or sizes. 
Hence, we standardize the puzzle format as multiple-choice questions, where all questions are provided with four options, with the exception of three options for size (small, medium, and large).
To generate the incorrect choices for each question, we use heuristics including randomly sampling numbers within the same magnitude of the answer, and further details can be found in the supplementary material.
We use the standard accuracy metric for evaluation.

\subsection{Dataset Analysis}
To ensure that the dataset contains diverse abstract patterns, we provide a taxonomy of 10 puzzle categories based on fundamental concepts including numbers, colors, shapes, and size.
As shown in Figure \ref{fig:ontology}, there are four categories of single-concept patterns.
To extend the depth of \datasetname{}, we also include dual-concept patterns, which would require models to relate two concepts in order to solve the puzzle.
Within each category, we design two multimodal templates that can each be used to generate many unique puzzle instances.
The full list of puzzle templates and examples of more puzzle instances can be found in the supplementary material.
To maintain a reasonable dataset size for evaluating large multimodal models, we generate 100 unique puzzle instances from each template. Thus, there are 2000 test instances in \datasetname{} in total.
We conducted an analysis in Appendix \ref{sec:data_size} and found that the chosen dataset size is large enough to be relatively robust to experimental variance.

\subsection{Implementation Details} We utilize Python code along with packages like Pillow\footnote{\url{https://pypi.org/project/pillow/}} to automatically generate puzzles. Leveraging these tools, we are able to create many different unique puzzle images and text questions for each given puzzle type by augmenting the base template and objects in the image. 
Example code snippets to generate the puzzles are included in the supplementary material, and we plan to release the dataset publicly with a permissive license such as MIT license.

Each puzzle in \datasetname{} comprises of an image $x_{image}$, a natural language question $x_{question}$, an image caption that describes the image $x_{caption}$,  an explanation that explains the pattern shown in the image $x_{explanation}$, a deduction statement that applies the pattern to the puzzle to derive the final answer $x_{deduction}$, a set of multiple-choice answers $x_{options}$, and the final answer $x_{answer}$. All of which are automatically generated during the puzzle creation process.

\begin{table}[!t]
\centering
\resizebox{\linewidth}{!}{
\begin{tabular}{lccccc}
\toprule
 & {Numbers} & {Colors} & {Size} & {Shapes} & {Average} \\
\midrule
Random Baseline & 25.0 & 25.0 & 33.3 & 25.0 & 27.1 \\
Qwen-VL-Chat & 25.0 & 22.0 & 32.5 & 34.0 & 28.4 \\
LLaVA-13B & 26.0 & 29.0 & 29.0 & 26.0 & 27.5 \\
Gemini Pro & 32.5 & 32.0 & 33.5 & 40.0 & 34.5 \\
Claude 3 Opus & 47.0 & 32.5 & 33.5 & \bf 44.5 & 39.4 \\
GPT-4V & \bf 67.5 & \bf 42.0 & \bf 35.0 & 41.0 & \bf 46.4 \\
\bottomrule
\end{tabular}
}
\caption{Accuracy of large multimodal models for single-concept abstract patterns in \datasetname{}.}
\label{tab:results_single}
\end{table}

\begin{table*}[!t]
\centering
\small
\resizebox{0.95\textwidth}{!}{
\begin{tabular}{lccccccc}
\toprule
 & Numbers & Colors & Numbers & Colors & Size & Colors & \\
 & \& & \& & \& & \& & \& & \& & Average \\
 & Shapes & Numbers & Size & Shapes & Shapes & Size & \\
\midrule
Random Baseline & 25.0 & 25.0 & 25.0 & 25.0 & 33.3 & 25.0 & 26.4 \\
Qwen-VL-Chat & 29.5 & 26.5 & 26.5 & 26.5 & 30.0 & 31.0 & 28.3 \\
LLaVA-13B & 20.0 & 30.0 & 31.5 & 36.0 & \bf 39.0 & 30.0 & 31.1 \\
Gemini Pro & 29.0 & 21.0 & 29.5 & 27.5 & 36.5 & 37.0 & 30.1 \\
Claude 3 Opus & 48.0 & 54.5 & \bf 34.0 & 42.5 & 33.0 & 50.0 & 43.7 \\
GPT-4V & \bf 52.5 & \bf 56.0 & 31.0 & \bf 47.0 & 31.5 & \bf 55.0 & \bf 45.5 \\
\bottomrule
\end{tabular}
}
\caption{Accuracy of large multimodal models for dual-concept abstract patterns in \datasetname{}. 
}
\label{tab:results_dual}
\end{table*}

\section{Experimental Setup}


\subsection{Inference Pipeline}
To elicit reasoning steps from large multimodal models, we leverage zero-shot chain of thought (CoT) prompting \citep{NEURIPS2022_8bb0d291} with a prompt similar to \code{``Let's think step by step''}. As the model may not always follow the same multiple-choice answer format, we also employ a model-based answer extraction stage.
Detailed examples of the prompts can be found in the supplementary material. 
Please note that our main experimental setting used in Table \ref{tab:results_single} and \ref{tab:results_dual} involves only the questions and images as multimodal inputs.
On the other hand, we progressively provide additional ground-truth information such as image captions in Section \ref{sec:bottlenecks} to diagnose the multimodal reasoning bottlenecks.

\paragraph{Chain of Thought Prompting.} \label{reasoning extraction}
In the first prompting step, we construct a prompt $\hat{x}$ by modifying the question using a specific prompt template: "[$I$] Question: [$X$]. Options: [$O$]. Answer: [$T$]", where [$I$] is the input slot for $x_{image}$, [$X$] is the input slot for $x_{question}$, [$O$] is the input slot for $x_{options}$, and [$T$] is the input slot for the trigger sentence $t_{1}$. 
To elicit reasoning over the multimodal inputs, we use \code{``Let's describe the image first and think step by step''} as our trigger sentence $t_{1}$. 
This modified prompt $\hat{x}$ is then fed into a large multimodal model, and a greedy decoding strategy is utilized to generate the subsequent sentence $y_{1}$. If the letter answer can be extracted from $y_{1}$ with regular expressions, the prompting process terminates. However, if the letter answer cannot be extracted, we prompt the model itself to extract the answer.

\paragraph{Answer Extraction.} In the second prompting stage, we use the generated sentence $y_{1}$ along with the modified prompt $\hat{x}$ to extract the final answer. We concatenate three elements to form "[$\hat{X}$] [$Z$] [$A$]" where [$\hat{X}$] is the input slot for $\hat{x}$, [$Z$] is the input slot for $y_{1}$, and [$A$] is the trigger sentence $t_{2}$ to extract the final answer. We defined $t_{2}$ as \code{"Therefore, among (A) (B) (C) (D), the answer is:"} or \code{"Therefore, among (A) (B) (C), the answer is:"}, for puzzles with four and three multiple-choice questions respectively.

\subsection{Models}
To investigate the reasoning ability of large multimodal models, we select the best-performing open and closed-source models \cite{yue2023mmmu}:
\begin{enumerate}
    \item Qwen-VL-Chat (7B) \cite{bai2024qwenvl} is an open-source large multimodal model designed to perceive and understand both texts and images. We use the version with open model weights and default chat template.
    \item LLaVA-13B \cite{liu2023improvedllava} is an large multimodal model which is based on the popular LLaMA \cite{touvron2023llama} foundation language model. We use the model weights of the 1.5 version and default chat template.
    \item Gemini Pro \cite{geminipro} is a highly capable multimodal model released by Google, and we use their publicly available API to query the ``gemini-pro-vision'' version of the model.
    \item Claude 3 Opus\footnote{\url{https://www.anthropic.com/news/claude-3-family}} is released by Anthropic and the most highest-performing multimodal model in their model family. We use their publicly available API to query the ``claude-3-opus-20240229'' version of the model.
    \item GPT-4V \cite{gpt4vision} is released by OpenAI and widely regarded as the most capable multimodal model based on existing benchmarks \cite{ yue2023mmmu}. We use their publicly available API to query the ``gpt-4-vision-preview'' version of the model.
\end{enumerate}

\section{Results}
\label{sec:results}

We report the main evaluation results on single-concept and dual-concept puzzles in Table \ref{tab:results_single} and Table \ref{tab:results_dual} respectively.
The evaluation results for single-concept puzzles, as shown in \Cref{tab:results_single} reveal notable differences in performance among the open-source and closed-source models. 
GPT-4V stands out with the highest average score of 46.4, demonstrating superior abstract pattern reasoning on single-concept puzzles such as numbers, colors, and size. 
It particularly excels in the "Numbers" category with a score of 67.5, far surpassing other models, which may be due to its advantage in math reasoning tasks \cite{dawn}. 
Claude 3 Opus follows with an overall average of 39.4, showing its strength in the "Shapes" category with a top score of 44.5. 
The other models, including Gemini Pro and LLaVA-13B trail behind with averages of 34.5 and 27.5 respectively, performing similarly to the random baseline on several categories.

In the evaluation on dual-concept puzzles, as shown in \Cref{tab:results_dual}, GPT-4V stands out again with the highest average score of 45.5. 
It performed particularly well in categories such as "Colors \& Numbers" and "Colors \& Size" with a score of 56.0 and 55.0 respectively. 
Claude 3 Opus closely follows with an average of 43.7, showing strong performance in "Numbers \& Size" with the highest score of 34.0.
Interestingly, LLaVA-13B, despite its lower overall average of 31.1, scores the highest in the "Size \& Shapes" category at 39.0.
Gemini Pro, on the other hand, has a more balanced performance across categories but with a slightly lower overall average of 30.1.
Overall, we find that models perform similarly on average for single-concept and dual-concept patterns, which indicates that they also struggle with puzzles that require reasoning about multiple abstract concepts.

\begin{figure}[t!]
\centering
\resizebox{1.0\linewidth}{!}{
\begin{tikzpicture}
    \begin{axis}[
        width=6.5cm,
        height=5cm,
        ybar=0cm,
        bar width=5pt,
        ymax=180,
        ymin=0,
        ylabel={Accuracy},
        label style={font=\fontsize{8}{1}\selectfont},
        xtick = {1,2,3,4,5},
        xticklabels = {LLaVA-13B, Gemini Pro, Claude Opus, GPT-4V},
        xticklabel style = {font=\fontsize{7}{1}\selectfont},
        yticklabel style = {font=\fontsize{7}{1}\selectfont},
        xtick pos = left,
        ytick pos = left,
        ymajorgrids = true,
        ytick={0,50,100},
        grid style=dashed, 
        legend style={
            font=\fontsize{6.8}{1}\selectfont, 
            legend style={row sep=-0.1cm},
            at={(1,1)},
        },
        legend image code/.code={
          \draw[#1] (0cm,-0.1cm) rectangle (0.4cm,0.05cm);
        }, 
        legend cell align={left},
    ]
    \addplot coordinates {
    (1, 27) (2, 34) (3, 39) (4, 46)};
    \addlegendentry{Original prompt (Only question \& image)};
    \addplot coordinates {
    (1, 33) (2, 43) (3, 72) (4, 70)};
    \addlegendentry{w/ Guided perception};
    \addplot coordinates {
    (1, 41) (2, 55) (3, 98) (4, 97)};
    \addlegendentry{w/ Guided perception, induction};
    \addplot[fill=teal] coordinates {
    (1, 58) (2, 80) (3, 98) (4, 99)};
    \addlegendentry{w/ Guided perception, induction, deduction};
    \end{axis}
\end{tikzpicture}
\vspace{-2mm}
}
\caption{Analysis on multimodal reasoning bottlenecks. We progressively prompt models with ground-truth
explanations for visual perception, inductive reasoning, and deductive reasoning.}
\label{fig:breakdown}
\end{figure}
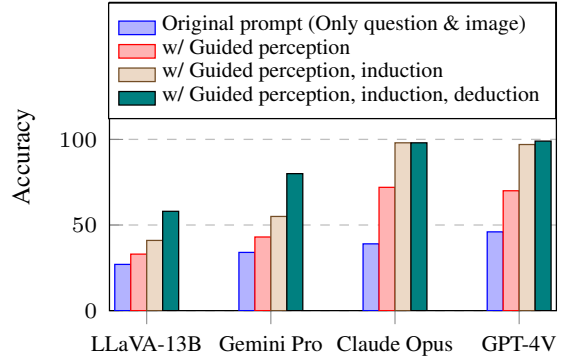

\subsection{Analysis of Multimodal Reasoning Bottlenecks}
\label{sec:bottlenecks}

Given the lower performance of existing large multimodal models, this raises the natural question of why they are not able to reason well about abstract patterns.
As shown in Figure \ref{fig:components}, the stages of solving abstract puzzles can be generally decomposed into visual perception, inductive reasoning, and deductive reasoning.
Hence, we analyze their reasoning bottlenecks in Figure \ref{fig:breakdown} by progressively providing the ground truth explanation in their prompts.
Note that we omit the final answer in the deductive reasoning explanation to avoid making the question trivial, and the detailed prompts can be found in the supplementary material.
Overall, we observe that the models perform better when provided with ground truth explanations, which suggests that they are able to leverage the additional information.

\ul{Notably, GPT-4V and Claude 3 Opus is able to solve almost all cases when provided with both visual perception and inductive reasoning explanations.
This suggests that the main bottlenecks for GPT-4V and Claude 3 Opus are visual perception and inductive reasoning to interpret the multimodal information and recognize the pattern from observations.}
However, this is not the case for LLaVA-13B and Gemini Pro, which demonstrate the largest improvement when guided by visual perception, inductive reasoning, and deductive reasoning together.
This indicates that their main bottleneck is deductive reasoning to apply general principles of the pattern to solve specific cases. 
Note that these results are intended to serve as an \textbf{optimistic upper bound} of the model performance when provided with ground truth partial information, and may not indicate that the puzzles will become trivial.

\begin{figure}[!t]
\centering
\resizebox{1.0\linewidth}{!}{
\begin{tikzpicture}
\begin{axis}[
    xbar,
    height=5cm,
    width=7cm,
    bar width=10pt,
    enlarge y limits={abs=14pt},
    symbolic y coords={Human Baseline, GPT-4V, Claude 3 Opus, Gemini Pro},
    nodes near coords,
    nodes near coords align={horizontal},
    xlabel={Accuracy},
    xmin=0,
    xmax=110,
    xtick={0,20,40,60,80,100},
]
\addplot[fill=blue!30] coordinates {
    (91.6,Human Baseline)
    (47.5,GPT-4V)
    (42.5,Claude 3 Opus)
    (27.5,Gemini Pro)
};
\end{axis}
\end{tikzpicture}
}
\caption{Comparison between average human performance and large multimodal models on a subset of \datasetname{}.}
\label{fig:human_baseline}
\end{figure}
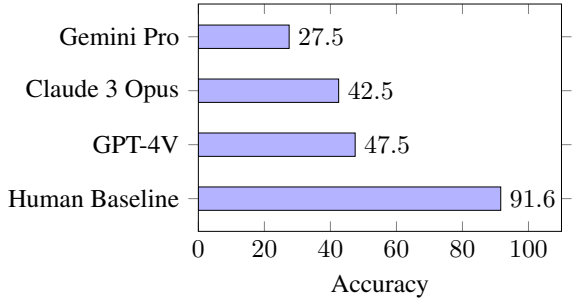


\subsection{Comparison to Human Performance}

To further shed light on how the large multimodal models compare to the reasoning ability of humans, we conducted a human baseline study involving 23 university students\footnote{Note that the participants volunteered for the short study and we obtained prior permission from their instructor.}. Participants were allotted 30 minutes to solve 40 puzzle instances sampled from our 20 puzzle categories, yielding an average human baseline score of 91.6\%, as shown in Figure \ref{fig:human_baseline}.
Note that the 20-24 age group of the participants correspond to the formal operational stage of cognitive development, as discussed in Section \ref{sec:cognitive}.
In contrast, the highest-performing model, GPT-4V scored 47.5\% on the same set of puzzle samples, highlighting the specific bottlenecks causing models to fall short of human cognition: primarily in visual perception and inductive reasoning, as discussed in Section \ref{sec:bottlenecks}. 


\begin{figure}
\centering
\resizebox{0.95\linewidth}{!}{
\begin{tikzpicture}
\pgfplotsset{width = 8cm, height = 5cm}
    \begin{axis}[
        xlabel={Number of Demonstrations},
        ylabel={Accuracy},
        ymax=52,
        ymin=16,
        xtick = {1,2,3,4,5,6,7,8,9},
        xticklabels = {0,1,2,3,4},
        xtick pos = left,
        ytick pos = left,
        legend pos = south east,
        legend style={font=\fontsize{8}{1}\selectfont, row sep=-0.1cm,/tikz/every odd column/.append style={column sep=0.01cm}},
        ymajorgrids = true,
        grid style=dashed,
    ]
    \addplot [mark=square, mark size=2pt, color=orange] plot coordinates {
    (1, 46.4) (2, 39.6) (3, 38.6) (4, 47.4) (5, 49.5)};
    \addlegendentry{GPT-4V};
    \addplot [mark=o,  mark size=2pt, color= blue] plot coordinates {
    (1, 31.5) (2, 35.5) (3, 35.4) (4, 34.2) (5, 37.2)};
    \addlegendentry{Gemini Pro};
    \addplot [mark=triangle,  mark size=2.5pt, color=teal] plot coordinates {
    (1, 27.3) (2, 29.0) (3, 28.4) (4, 29.9) (5, 32.1)};
    \addlegendentry{LLaVA-13B};
    \end{axis}
\end{tikzpicture}
}
\caption{Analysis on the effect of few-shot demonstrations on model performance for single-concept puzzles.}
\label{fig:few_shot}
\end{figure}
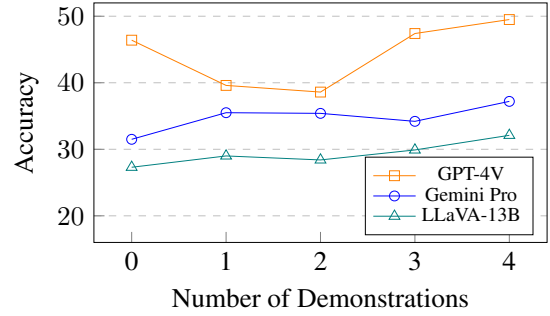

\subsection{Effect of Few-Shot Demonstrations}
\label{sec:few_shot}

While we focus on the zero-shot setting to investigate how multimodal models handle novel reasoning challenges, we also explore how models may use knowledge and strategies from other puzzles to solve a new, specific puzzle.
This is akin to analogical reasoning, which involves using experience from similar scenarios to make inferences about a novel situation \cite{emergentanalogical}.
Concretely, we run an analysis to study the effect of in-context learning \cite{gpt3} with few-shot demonstrations of other puzzle instances when the model is tasked to solve a specific puzzle. The demonstrations consist of interleaved instances of multimodal inputs of each puzzle image and question, as well as the ground truth reasoning explanations. 
To ensure that the demonstrations are sufficiently diverse, we randomly select puzzles of different categories from the given puzzle.
As shown in Figure \ref{fig:few_shot}, we find a general trend of increasing performance with respect to the number of demonstrations.
Although there are some cases of lower performance for GPT-4V, we see that models generally achieve their best performance with the most number of demonstrations.
This suggests that the models are indeed capable of analogical reasoning, and in-context learning may be a promising direction to enhance the abstract reasoning abilities of multimodal models in the future.

In addition, while not the main focus of this work, there may be other methods of improving model performance, including model training or different prompting methods.
Hence, we have also included preliminary studies on fine-tuning with LLaVA-13B and comparison between chain-of-thought and direct prompting in the supplementary material.
To consider the effect of other factors, the supplementary material further includes an alternative setting with text-only models given the ground-truth visual perception caption, and an analysis on the effect of evaluation dataset size.

\section{Related Work}


The recent surge in multimodal pretraining and fine-tuning approaches \cite{liu2023improvedllava, bai2024qwenvl} has led to the creation of various benchmarks. Benchmarks like VQA \cite{VQA} and OK-VQA \cite{okvqa} aim at evaluating the basic perception and reasoning abilities of large multimodal models. Meanwhile, benchmarks like MMMU \cite{yue2023mmmu} and ScienceQA \cite{scienceqa} offer an evaluation of LLMs' proficiency across multiple disciplines requiring domain-specific knowledge and multimodal understanding.

To investigate the fundamental challenges in multimodal perception and reasoning, we deliberately focus on the abstract domain, aiming to assess how models emulate cognitive abilities, particularly involving reasoning about abstract concepts and relationships. 
However, we note that existing benchmarks have limitations which make them less suitable for studying large multimodal models. The RAVEN dataset \cite{zhang2019raven} presents visual matrices with abstract patterns, challenging models to identify patterns and complete missing elements.
However, we note that it has a specific spatial layout and can be solved exactly with search algorithms.
Compared to CLEVR \cite{johnson2017clevr} which offers synthetic visual scenarios and questions focusing on logic and commonsense, we focus on exploring how large multimodal models perceive and reason about multimodal patterns, which is more closely related to fundamental cognitive processes in humans \cite{Mattson2014SuperiorPP}. 
While ConceptARC \cite{conceptarc} focuses on specific spatial concepts such as inside-outside and above-below, our dataset \datasetname{} studies how visual objects interact and relate based on broader abstract concepts such as colors, shapes, numbers, and size.
Lastly, the MiniSCAN dataset \cite{miniscan} presents patterns that map special words to a sequence of color symbols, but is limited to color-based patterns.

What distinguishes our dataset, \datasetname{}, from the existing works is its systematic analysis of multimodal reasoning through abstract patterns, including perceptual, inductive, and deductive reasoning.
Compared to the previous datasets, our multimodal patterns encompass broad and fundamental abstract concepts such as numbers, colors, shapes, and size.
Notably, our dataset not only provides ground truth answers but also includes image captions and pattern explanations that enable more detailed and systematic diagnosis of the reasoning bottlenecks for large multimodal models.

\section{Conclusion}
In this work, we introduced the \datasetname{} dataset to investigate the reasoning challenges in large multimodal models.
Our experiments demonstrated that, despite their sophistication, models such as GPT-4V exhibit substantial challenges when solving abstract pattern puzzles that require visual perception, inductive reasoning, and deductive reasoning, falling short of cognitive processes displayed by humans.
Notably, our systematic analysis with ground truth explanations reveals that the main reasoning bottlenecks for GPT-4V are weaker visual perception and inductive reasoning capabilities. 
On the other hand, we found that other large multimodal models required more guidance with ground truth explanations, pointing to a broader range of reasoning challenges.
Looking ahead, our work points to exciting avenues for advancing the reasoning abilities of large multimodal models. 
Future research should focus on enhancing models' understanding of multimodal information and refining their abstract reasoning faculties, in order to further enhance their general capabilities.


\section*{Acknowledgement}
This work was substantially supported by DAMO Academy through DAMO Academy Research Intern Program. This work was partially supported by AI Singapore Governance grant ID: AISG3-GV-2023-010, and AcRF MoE Tier-2 grant (Project no. T2MOE2008, Grantor reference no. MOE-T2EP20220-0017) titled: “CSK NLP: Leveraging Commonsense Knowledge for NLP”.
Chia Yew Ken would like to thank Tan Hui Min Grace as a source of inspiration for fun and unique puzzles.

\clearpage
\newpage

\section*{Limitations}
In this work, we mainly focus on the zero-shot setting to investigate how large multimodal models face reasoning challenges in novel situations.
However, previous works have shown that prompting with demonstrations \cite{gpt3} may improve the models ability to adapt to new tasks. 
Hence, we also include experiments in the few-shot setting in Section \ref{sec:few_shot}, which showed inconsistent benefits, and we aim to explore this area in the future works.



\bibliography{custom}

\clearpage
\newpage
\appendix

\section{Appendix}
\label{sec:appendix}

\subsection{Multiple-Choice Format Details}
\label{sec:multiple_choice}
To generate multiple choice options for numeric puzzles, we use heuristics based on the range of the number. For example, if the number is less than 10, then we sample from the range 1 to 9. If the number if less than 100, we sample from the range 1 to 99, and so on.
For discrete option choices, we sample from the possible objects in the image, such as the list of predefined colors or sizes or shapes.

\subsection{Dataset Details}
\label{sec:data_details}

We report the dataset statistics of \datasetname{} in Table \ref{tab:dataset}.

\begin{table}[!t]
\centering
\resizebox{\linewidth}{!}{
\begin{tabular}{lccccc}
\toprule
Puzzle & Multimodal & Test \\
Category & Templates & Instances \\
\midrule
Numbers & 2 & 200 \\
Colors & 2 & 200 \\
Shapes & 2 & 200 \\
Size & 2 & 200 \\
Numbers \& Shapes & 2 & 200 \\
Numbers \& Colors & 2 & 200 \\
Numbers \& Size & 2 & 200 \\
Shapes \& Colors & 2 & 200 \\
Shapes \& Size & 2 & 200 \\
Colors \& Size & 2 & 200 \\
\midrule
Total & 20 & 2000 \\
\bottomrule
\end{tabular}
}
\caption{Dataset statistics of \datasetname{}.}
\label{tab:dataset}
\end{table}

\subsection{Prompt Examples}
\label{sec:prompts}

\begin{figure*}[t]
    \centering
    \includegraphics[width=\textwidth]{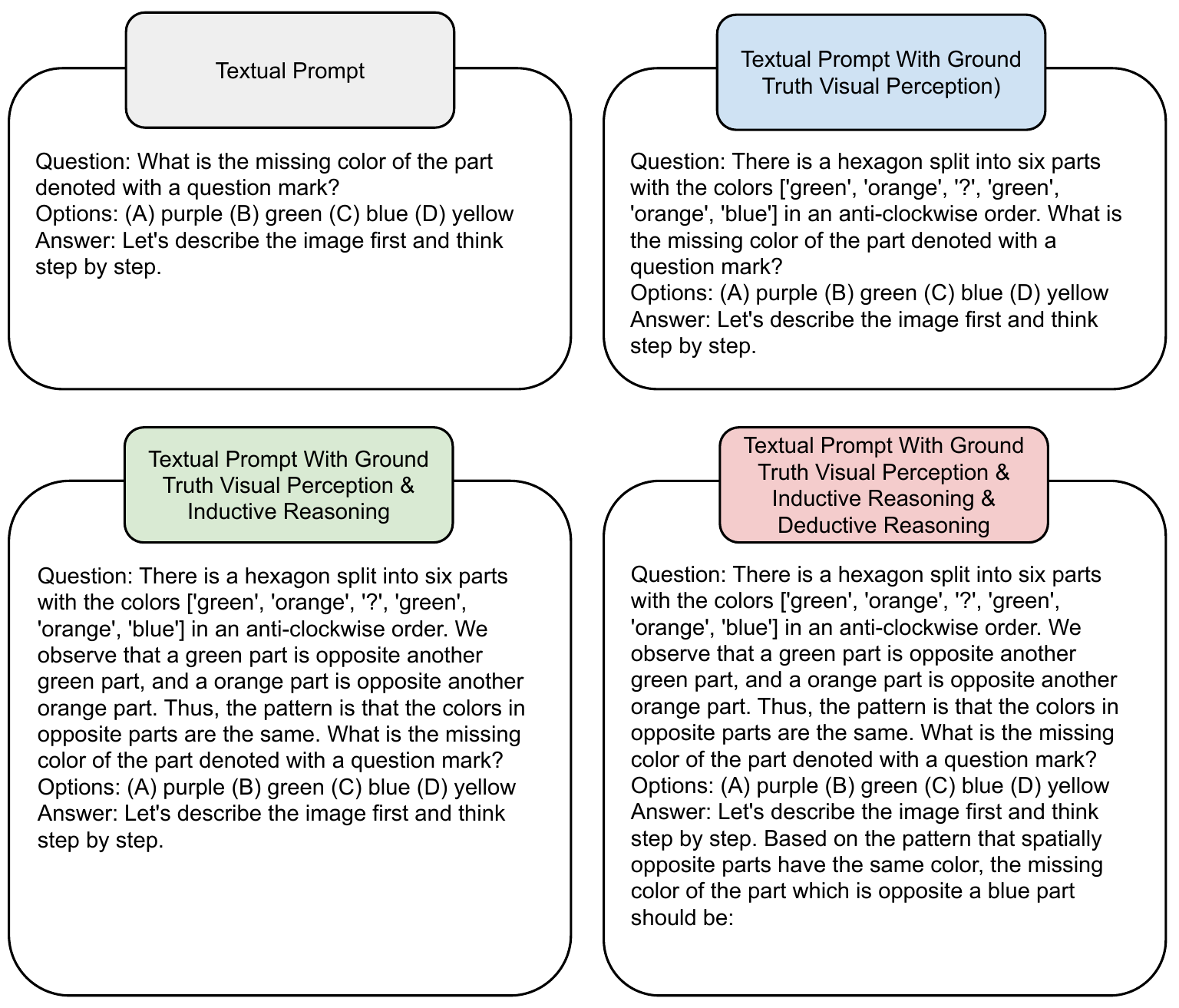}
    \caption{Textual prompt examples for eliciting reasoning steps from large multimodal models.}
    \label{fig:prompts}
\end{figure*}

We show examples of the textual prompts in Figure \ref{fig:prompts}. Note that the prompt examples correspond to the image and puzzle in Figure \ref{fig:intro}.
We use a consistent prompt format across all abstract puzzles.

\subsection{Code Implementation Example}
\label{sec:code_snippet}

\lstset{
    language=Python,
    basicstyle=\ttfamily\tiny,
    keywordstyle=\color{blue},
    stringstyle=\color{red},
    commentstyle=\color{green},
    morecomment=[l][\color{magenta}]{\#},
    frame=single,
    breaklines=true,
    postbreak=\mbox{\textcolor{red}{$\hookrightarrow$}\space},
    showstringspaces=false,
    tabsize=4,
    backgroundcolor=\color{gray!10}
}

\begin{lstlisting}
import math
import random
from typing import List, Tuple, Dict

from PIL import Image, ImageDraw, ImageFont
from pydantic import BaseModel

class ColorHexagonPattern(BaseModel):
    colors: Dict[str, str] = dict(
        blue="#6fa8dc",
        green="#93c47d",
        yellow="#ffd966",
        red="#e06666",
        purple="#8e7cc3",
        orange="#f6b26b",
    )
    image_size: int = 512
    scale_factor: int = 4
    path_font: str = "fonts/OpenSans-Medium.ttf"

    @staticmethod
    def get_centroid(points: List[Tuple[float, float]]) -> Tuple[float, float]:
        x = sum(p[0] for p in points) / len(points)
        y = sum(p[1] for p in points) / len(points)
        return x, y

    def sample_colors(self) -> Tuple[List[str], List[str]]:
        while True:
            names = random.sample(list(self.colors), k=3)
            if "orange" in names and "yellow" in names:
                continue  # Hard to distinguish
            names = names + names
            colors = [self.colors[n] for n in names]
            return names, colors

    def make_sample(self):
        # Set the size of the image
        size = self.image_size * self.scale_factor
        image = Image.new("RGB", size=(size, size), color="white")
        draw = ImageDraw.Draw(image)
        center = size // 2

        # Hexagon properties
        length = size // 3  # Length of a side of the hexagon and triangles
        triangle_height = math.sqrt(3) / 2 * length

        # The vertices of the hexagon
        hexagon = [
            (center + length / 2, center - triangle_height),
            (center - length / 2, center - triangle_height),
            (center - length, center),
            (center - length / 2, center + triangle_height),
            (center + length / 2, center + triangle_height),
            (center + length, center),
        ]

        # Colors for the triangles
        names, colors = self.sample_colors()
        i_answer = random.randint(0, len(colors) - 1)
        answer = names[i_answer]
        colors[i_answer] = "#eeeeee"  # Grey

        # Draw the hexagon made of six triangles
        for i in range(6):
            # Coordinates of the triangle vertices
            triangle = [hexagon[i], hexagon[(i + 1) % 6], (center, center)]
            # Draw the triangle
            draw.polygon(triangle, fill=colors[i])
            # Draw the outline with custom width
            points = [hexagon[i], hexagon[(i + 1) % 6], (center, center), hexagon[i]]
            draw.line(points, fill="black", width=self.scale_factor * 4)
            # Draw "?" on the missing answer part
            if i == i_answer:
                draw.text(self.get_centroid(triangle),
                    text="?",
                    font=ImageFont.truetype(self.path_font, size=size // 10),
                    anchor="mm",
                    fill="black",
                )

        names[i_answer] = "?"
        instances = sorted(set(n for n in names if n not in [answer, "?"]))
        image = image.resize((self.image_size, self.image_size), Image.LANCZOS)
        return (
            dict(
                question="What is the missing color of the part denoted with a question mark?",
                answer=answer,
                options=sample_options(answer, options=list(self.colors), k=4),
                caption=f"There is a hexagon split into six parts with the colors {names} in an anti-clockwise order.",
                explanation=f"We observe that a {instances[0]} part is opposite another {instances[0]} part, and a {instances[1]} part is opposite another {instances[1]} part. Thus, the pattern is that the colors in opposite parts are the same.",
                deduction=f"Based on the pattern that spatially opposite parts have the same color, the missing color of the part which is opposite a {answer} part should be {answer}.",
            ),
            image,
        )
\end{lstlisting}

\subsection{Dataset Size Analysis}
\label{sec:data_size}

Regarding the dataset size and diversity, we set the number of generated instances for each puzzle to 100, to reduce experimental variance and maintain a reasonable evaluation cost. Hence, the current dataset size is 2000 samples (20 templates with 100 instances each). As there are two templates per puzzle category, this means there are 200 test samples for each puzzle category. To observe the impact of the number of test samples, we evaluate the models on three different data settings as shown in Table \ref{tab:sizes}: 50, 100, and 200 test samples in per puzzle, which correspond to 1000, 2000, and 4000 total samples respectively. In general, we observe some variations in the average score for single-concept puzzles, but it does not significantly affect the comparison of performance between different models. Hence, we believe that the chosen dataset size is large enough to be relatively robust to experimental variance. To investigate the multimodal reasoning capabilities across diverse abstract scenarios, we construct the puzzles based on four fundamental concepts: numbers, colors, shapes, and size. Furthermore, to evaluate how well the models can relate to multiple concepts, we design both single-concept and dual-concept puzzles, and the taxonomy of diverse puzzles is shown in Figure \ref{fig:ontology}.

\begin{table}[!t]
\centering
\resizebox{\linewidth}{!}{
\begin{tabular}{lccccc}
\toprule
& 1000 & 2000 & 4000 \\
\midrule
Gemini Pro Avg. Score & 29.2 & 34.5 & 32.2 \\
GPT-4V Avg. Score & 48.8 & 46.4 & 48.4 \\
\bottomrule
\end{tabular}
}
\caption{Analysis of model performance with respect to number of testing data samples.}
\label{tab:sizes}
\end{table}

\subsection{Comparison of CoT and Direct Prompting}
\label{sec:direct_prompt}

To investigate the effect of direct prompting without chain of thought, we evaluated the models on our main setting as shown in Table \ref{tab:results_single} and \ref{tab:results_dual}. The results are shown in Table \ref{tab:direct_prompt_results} for single-concept puzzles and indicate that direct prompting is less effective for Gemini Pro and GPT-4V models, compared to CoT prompting in Table \ref{tab:results_single}. This may be due to differences in training data and alignment methods between the models.

\begin{table}[!t]
\centering
\resizebox{\linewidth}{!}{
\begin{tabular}{lccccc}
\toprule
& Numbers & Colors & Size & Shapes & Average \\
\midrule
LLaVA-13B & 25.5 & 29.0 & 37.0 & 39.0 & 32.6 (+5.1) \\
Gemini Pro & 16.0 & 28.5 & 35.0 & 49.0 & 32.1 (-2.4) \\
GPT-4V & 72.5 & 26.5 & 30.0 & 45.5 & 43.6 (-2.8) \\
\bottomrule
\end{tabular}
}
\caption{Results of direct prompting and change in average performance compared to CoT prompting.}
\label{tab:direct_prompt_results}
\end{table}

\begin{figure*}[t]
    \centering
    \includegraphics[width=\textwidth]{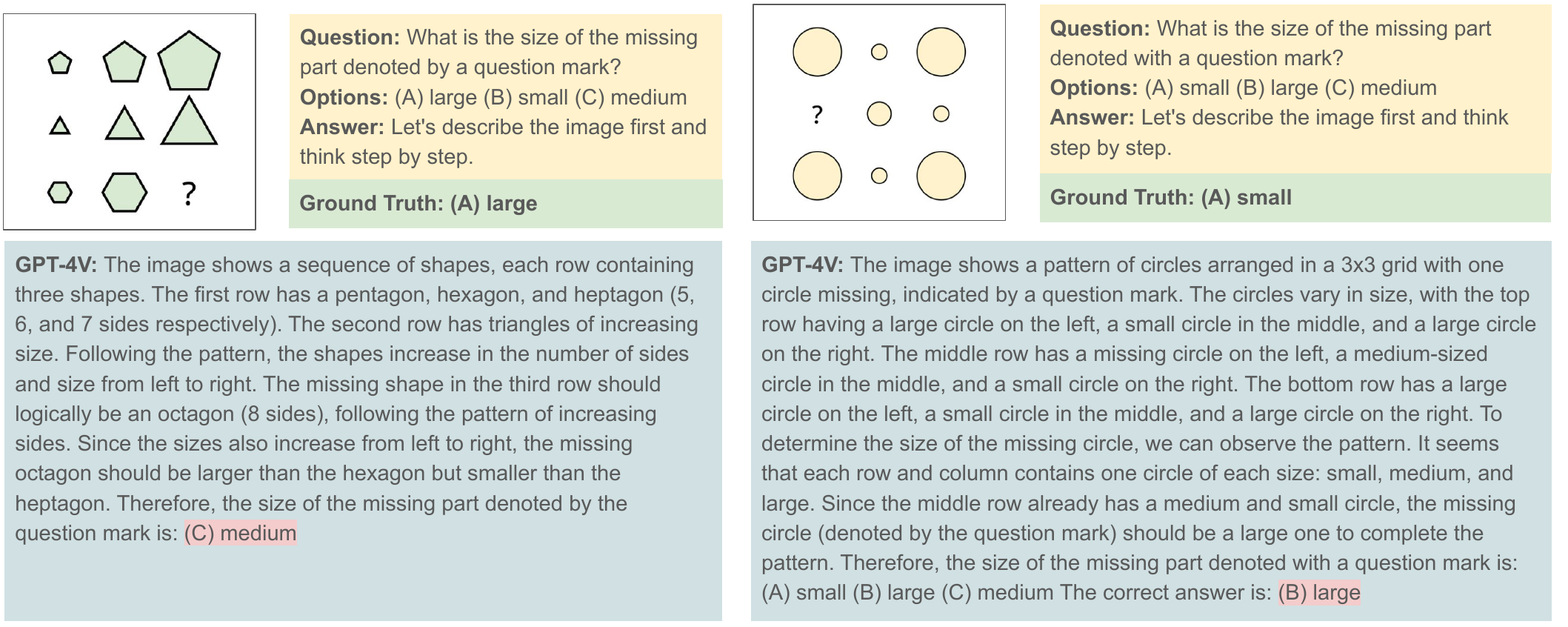}
    \caption{Case study on two sample predictions from GPT-4V. The example on the left shows visual perception failures and the example on the right shows the faulty inductive reasoning of the model which proposed a spurious pattern in the image.}
    \label{fig:case}
\end{figure*}

\section{Qualitative Analysis}

To illustrate the reasoning bottlenecks of GPT-4V, we include two case study samples in Figure \ref{fig:case}.
For instance, the sample on the left is from the size \& shapes category of puzzles, for which the model under-performed the random baseline.
For visual perception, we observe that the model presents severe limitations, as it is unable to recognize simple polygon shapes and hallucinated additional shapes which are not in the image.
Regarding inductive reasoning, we observe that the model was able to recognize the sizes of the different objects, but did not recognize the correct pattern that the circles directly adjacent to the center should be small in size.
Hence, we believe that there is ample area for improvement for abstract reasoning ability in large multimodal models.

\end{document}